\documentclass[sigconf]{acmart}
\usepackage{balance}
\usepackage{hyperref}
\usepackage{cleveref}
\usepackage{bm}
\usepackage{subcaption}
\usepackage{makecell} 
\usepackage{amsmath}
\usepackage{cuted}
\AtBeginDocument{%
  }
    
\copyrightyear{2025}
\acmYear{2025}
\setcopyright{acmlicensed}
\acmConference[MMAsia '25]{ACM Multimedia Asia}{December 9--12, 2025}{Kuala Lumpur, Malaysia}
\acmBooktitle{ACM Multimedia Asia (MMAsia '25), December 9--12, 2025, Kuala Lumpur, Malaysia}
\acmDOI{10.1145/3743093.3771049}
\acmISBN{979-8-4007-2005-5/2025/12}

\acmSubmissionID{393}              

\begin{document}

\title{Sketch-1-to-3: One Single Sketch to 3D Detailed Face Reconstruction}
\author{Liting Wen}
\authornote{Both authors contributed equally to this research.}
\orcid{0000-0001-9833-8496} 
\affiliation{%
  \institution{Carnegie Mellon University}
  \city{Pittsburgh}
  \state{Pennsylvania}
  \country{USA}
}
\email{litingw@andrew.cmu.edu}

\author{Zimo Yang}
\authornotemark[1]
\orcid{0009-0003-2057-3949} 
\affiliation{%
  \institution{Nanyang Technological University}
  \city{Singapore}
  \country{Singapore}
}
\email{zimo002@e.ntu.edu.sg}

\author{Xianlin Zhang}
\authornote{Corresponding author.}
\orcid{0000-0003-3905-2062} 
\affiliation{%
  \institution{Beijing University of Posts and Telecommunications}
  \city{Beijing}
  \country{China}}
\email{zxlin@bupt.edu.cn}

\author{Chi Ding}
\orcid{0009-0001-2873-0705} 
\affiliation{%
  \institution{Beijing University of Posts and Telecommunications}
  \city{Beijing}
  \country{China}
}
\email{charmingchi@bupt.edu.cn}

\author{Mingdao Wang}
\orcid{0009-0004-2907-086X} 
\affiliation{%
 \institution{Tsinghua University}
 \city{Beijing}
 \country{China}}
\email{wmingdao@tsinghua.edu.cn}

\author{Xueming Li}
\orcid{0000-0003-1058-2799} 
\affiliation{%
  \institution{Beijing University of Posts and Telecommunications}
  \city{Beijing}
  \country{China}}
\email{lixm@bupt.edu.cn}
\renewcommand{\shortauthors}{Wen et al.}
\begin{abstract}
3D face reconstruction from a single sketch is a critical yet underexplored task with significant practical applications. The primary challenges stem from the substantial modality gap between 2D sketches and 3D facial structures, including: (1) accurately extracting facial keypoints from 2D sketches; (2) preserving diverse facial expressions and fine-grained texture details; and (3) training a high-performing model with limited data. In this paper, we propose Sketch-1-to-3, a novel framework for realistic 3D face reconstruction from a single sketch, to address these challenges. Specifically, we first introduce the \textbf{G}eometric \textbf{C}ontour and \textbf{T}exture \textbf{D}etail (GCTD) module, which enhances the extraction of geometric contours and texture details from facial sketches. Additionally, we design a deep learning architecture with a domain adaptation module and a tailored loss function to align sketches with the 3D facial space, enabling high-fidelity expression and texture reconstruction. To facilitate evaluation and further research, we construct SketchFaces, a real hand-drawn facial sketch dataset, and Syn-SketchFaces, a synthetic facial sketch dataset. Extensive experiments demonstrate that Sketch-1-to-3 achieves state-of-the-art performance in sketch-based 3D face reconstruction.
\end{abstract}


\maketitle

\begin{figure}[!t]
    \centering
    \includegraphics[width=0.99\linewidth]{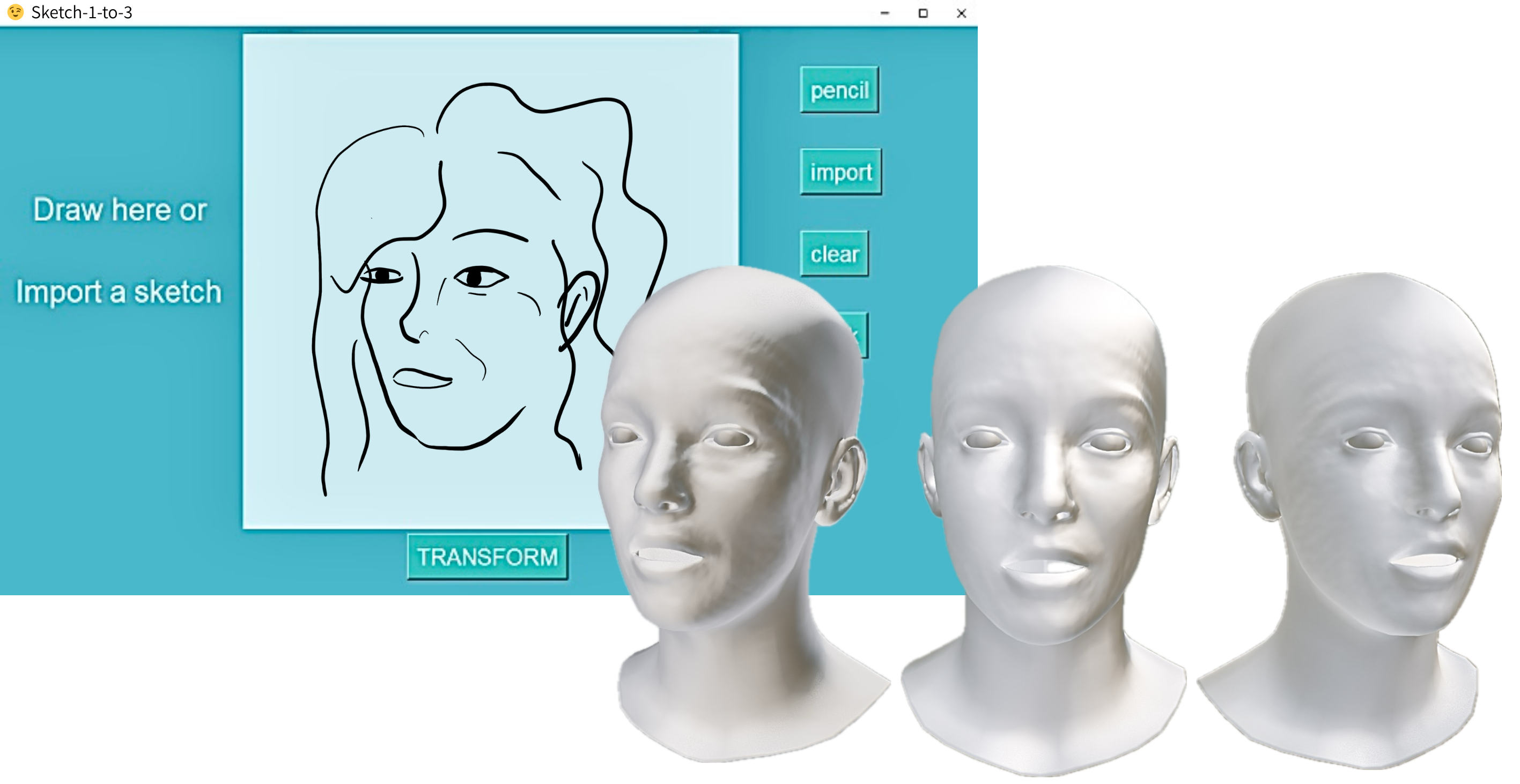}
    \caption{A user draws a facial sketch (left), and our Sketch-1-to-3 system produces a detailed 3D face reconstruction (right) that faithfully preserves geometry and fine details, remaining robust to occlusions and stylistic stroke variations.}
    \Description{The figure shows an interface of the Sketch-1-to-3 system on the left, where a user has drawn a simple free-hand facial sketch with black lines on a light blue background. To the right of the interface are three rendered views of the reconstructed 3D face model: a left-facing view, a frontal view, and a right-facing view. The reconstructed face is realistic, preserving the geometry and fine details implied by the sketch.}
    \label{fig:teaser}
    \vspace{-6pt}
\end{figure}
\section{Introduction}
3D face reconstruction from a single sketch has attracted considerable attention due to its wide-ranging applications, including animation creation~\cite{wu2023audio}, face recognition~\cite{wang2025implicit}, model retrieval~\cite{yang2024sketch}, and facial texture generation~\cite{wang2024s2td}. Although many existing works have focused on reconstructing 3D faces from richly informative images, the task of reconstructing high-fidelity 3D faces from a single hand-drawn sketch remains significantly more challenging. Unlike real images, sketches inherently contain sparse information, abstract representations, and user-specific stylistic variances, making accurate 3D reconstruction particularly difficult. Achieving a satisfying reconstruction from a single sketch involves faithfully capturing intended geometric contours and subtle facial details without introducing misleading embellishments or omissions.

\begin{figure}[t]
    \centering
    \includegraphics[width=\linewidth]{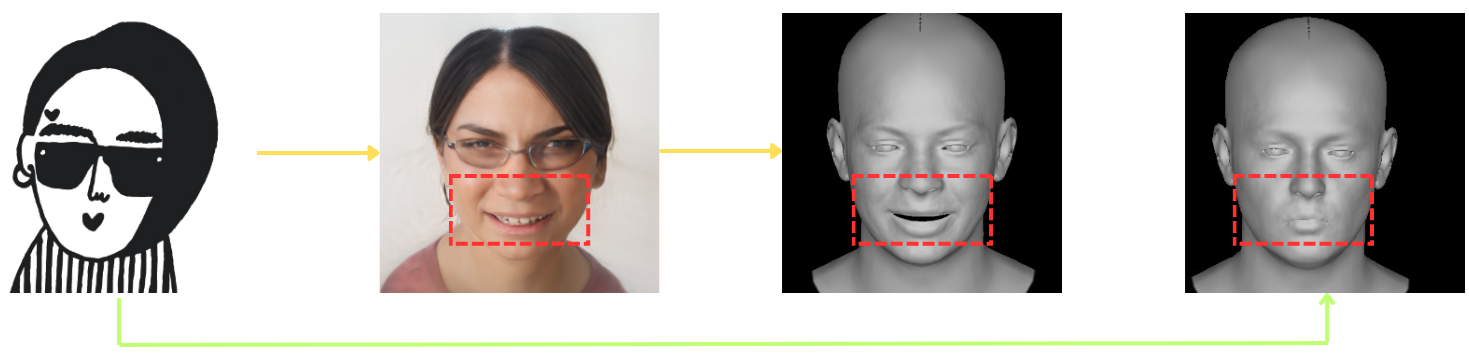}
    \caption{Challenge demonstration. The yellow line denotes a sketch-photo-3D pipeline using \cite{lin2023sketchfacenerf} and \cite{feng2021learning}, while the green line represents our method. The indirect sketch-photo-3D process tends to introduce reconstruction bias.}
    \Description{A diagram comparing two pipelines for 3D face reconstruction from sketches. The yellow line represents an indirect method that first converts the sketch into a photo and then reconstructs the 3D face from the photo. The green line represents our direct method. The comparison illustrates that the indirect method may introduce reconstruction bias.}
    \label{fig:sketch-2D-3D}
    \vspace{-20pt}
\end{figure}

Specifically, despite significant advancements in separate sketch-to-image and image-to-3D face reconstruction methods, we observe that naively combining these two methods for sketch-to-3D face reconstruction introduces a significant amount of misleading information that is not inherently associated with the original sketch, thus impeding a faithful reconstruction of the user's original intent, as illustrated in \hyperref[fig:sketch-2D-3D]{Figure~\ref{fig:sketch-2D-3D}}. This observation highlights a critical limitation of existing pipelines: they are not designed to handle the unique challenges of directly reconstructing 3D faces from sketches, where the input is sparse, abstract, and rich in user-specific stylistic variances. Therefore, a dedicated sketch-to-3D face reconstruction framework is desirable and essential. In this paper, we present a novel method specifically tailored to minimize misleading artifacts while preserving essential geometric structures and facial details, enabling faithful recovery of the user’s original intent.

We identify key limitations in sketch-based 3D face reconstruction, including the inherent ambiguity of hand-drawn sketches that hinders reliable information extraction, the difficulty of reconstructing high-fidelity 3D faces with vivid expressions and details from abstract line drawings, and the scarcity of high-quality facial sketch datasets for training. To overcome these challenges, we introduce Sketch-1-to-3, a novel framework that reconstructs high-fidelity 3D faces from a single sketch by accurately extracting contours and facial details. Furthermore, we contribute two datasets, SketchFaces and Syn-SketchFaces, to address the data scarcity problem.

In summary, our main contributions are:
\begin{itemize}
\item We propose an end-to-end framework, Sketch-1-to-3, for high-fidelity 3D face reconstruction from a single sketch. Our method is the first to explicitly address the precise transmission of information within a single sketch, enabling the reconstruction of a 3D face model with both accurate contours and fine-grained details.

\item We introduce a novel enhancement \textbf{G}eometric \textbf{C}ontour and \textbf{T}exture \textbf{D}etail (GCTD) module, which effectively extracts accurate geometric structures and fine details from sketches. We further design a domain adaption module and a task-specific loss function to seamlessly align 2D sketches with 3D space, ensuring expressive and detail-preserving 3D face reconstruction.

\item To address the critical data scarcity challenge in this domain, we introduce two comprehensive datasets: SketchFaces, containing real hand-drawn facial sketches, and Syn-SketchFaces, a large-scale synthetic sketch face dataset. These datasets can significantly advance research in sketch-based 3D face learning.

\item Extensive experiments show that Sketch-1-to-3 achieves state-of-the-art result in sketch-based 3D face reconstruction compared with previous methods.

\end{itemize}

\section{Related work}

\subsection{Monocular 3D Face Reconstruction} 
Reconstructing 3D faces from a single 2D image is an inherently ill-posed problem, requiring supplementary prior knowledge to constrain the reconstruction process. A widely adopted approach is to estimate the parameters of a 3D Morphable Model \cite{blanz2003face}. Existing methods can be broadly categorized into optimization-based \cite{aldrian2012inverse,thies2016face2face,zhu2015high,huber2015fitting,ploumpis2020towards} and learning-based techniques \cite{jourabloo2016large,richardson2017learning,deng2019accurate,sanyal2019learning,feng2021learning}. 
A major challenge for learning-based methods is the lack of ground-truth 2D–3D paired data. Inspired by \cite{sela2017unrestricted}, which shows that accurate 3D face reconstruction is possible without 3D supervision using synthetic data, we propose a sketch-based method that reconstructs detailed 3D faces from a single sketch using only 2D supervision on real and synthetic data.
\subsection{Sketch-based 2D Face Generation} 
There has been significant progress in converting free-hand sketches into realistic 2D face images. To bridge cross-modal gaps in sketch-to-photo synthesis, GLAS~\cite{li2024few} adopts a global-local fusion network within a few-shot asymmetric translation framework. SC-FEGAN \cite{Alpher10} generates images when users provide free-form masks, sketches, and colors as input. DrawingInStyles~\cite{Alpher14} incorporates spatial constraints from sketches and semantic maps into StyleGAN~\cite{karras2019style} to achieve high-quality and editable 2D face generation. DeepFacePencil~\cite{Alpher16} employs spatial attention pooling and a dual-generator design to adaptively reconcile the domain differences between sketches and face images, allowing generalization to freehand inputs. While these methods effectively translate facial sketches into 2D images using multimodal strategies, they remain limited to 2D generation without addressing detailed 3D face reconstruction. Building on these insights, we take a step further by reconstructing a 3D face from a single sketch.
\subsection{Sketch-based 3D Face Reconstruction} 
Although 2D sketch-to-face generation has advanced considerably, reconstructing 3D faces directly from sketches remains underexplored. DeepSketch2Face~\cite{han2017deepsketch2face} introduces a deep learning-based system that reconstructs 3D caricature-style faces from sketches. While it enables rapid generation of facial geometry and expressions from free-hand inputs, the results are often cartoonish and lack fine-grained details, limiting its applicability to high-fidelity 3D face reconstruction. SketchFaceNeRF~\cite{lin2023sketchfacenerf} requires significant training time and fails to recover complete geometry from a single sketch, yielding only partial-view models. A subsequent work~\cite{Alpher22} proposes a two-stage method that first combines sketches and reference photos for coarse reconstruction, then refines the 3D face with sketch contours. However, the results still fail to produce detailed 3D faces. Motivated by these limitations, we propose Sketch-1-to-3, a novel framework for reconstructing the high-fidelity 3D face from a single sketch by explicitly capturing and transmitting structural cues to recover both accurate contours and fine-grained details.
\section{Method}
\begin{figure*}[t]
	\centering
	\includegraphics[width=0.9\linewidth]{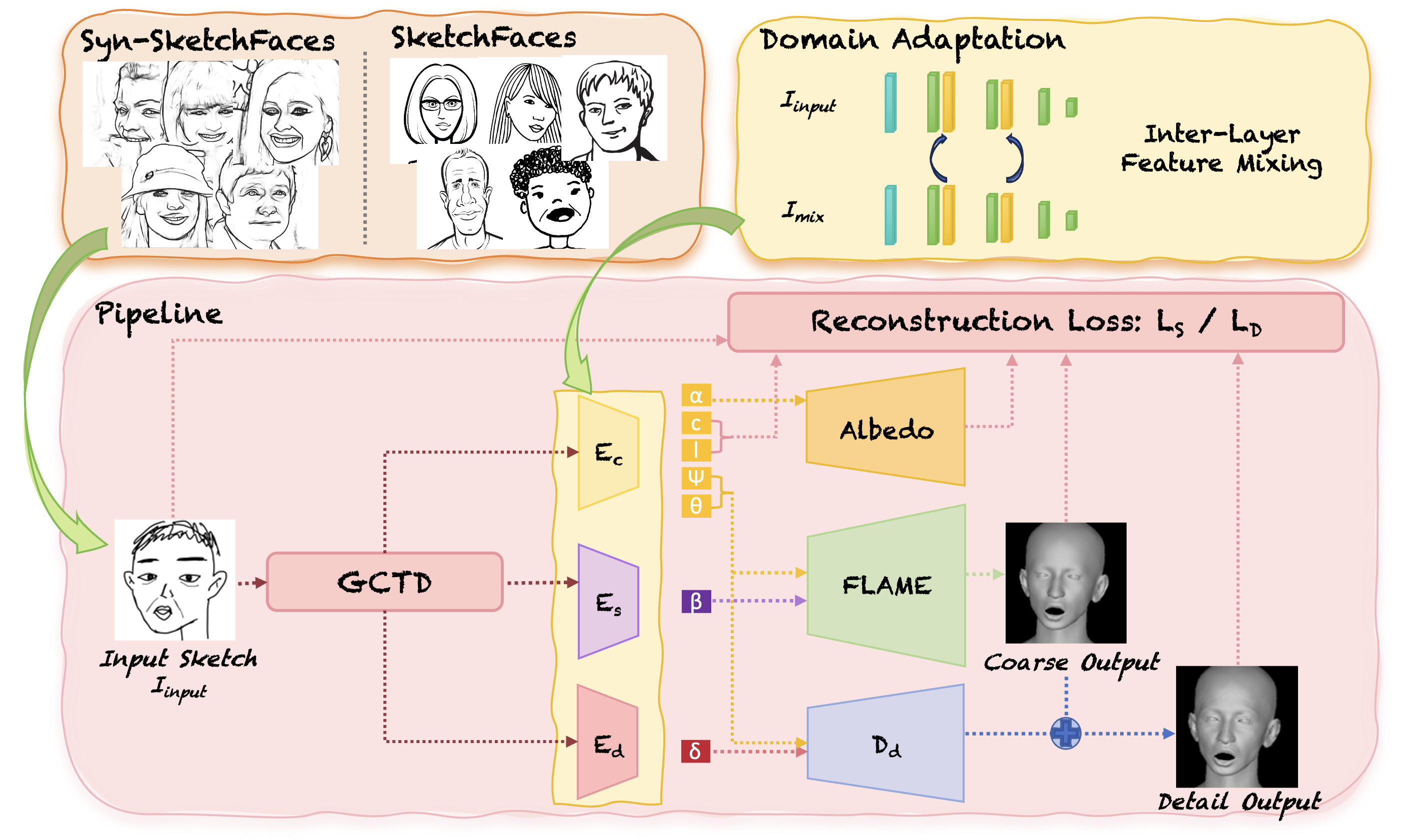}
	\caption{An overview of the proposed method, which consists of coarse and detail training stages. In the coarse stage, a shape encoder \(E_{s}\) regresses \(\bm{\beta}\) and a coarse encoder \(E_{c}\) regresses \(\bm{\theta}\), \(\bm{\psi}\), \(\bm{l}\), \(\bm{c}\), and \(\bm{\alpha}\). In the detail stage, a detail encoder \(E_{d}\) generates a latent code \(\bm{\delta}\) to refine the coarse 3D face with fine details. The 3D faces from both stages are rendered into 2D and compared with the input sketch to compute reconstruction losses. A GCTD module is applied in both stages to enhance feature extraction. All encoders employ inter-layer feature mixing for domain adaptation.}
    \Description{An overview of the proposed method.}
	\label{fig:framework}
    \vspace{-6pt}
\end{figure*}

Our goal is to reconstruct the 3D detailed face from a single sketch. Face sketches often exhibit two traits: (1) unrestrained composition that may deviate from real-face symmetry, necessitating adaptability to varying styles and distortions; and (2) strokes that mix noise with critical intentions from the creator, highlighting the need for noise suppression and critical feature enhancement. Motivated by these observations, we propose a sketch-to-3D detailed face reconstruction method with a novel feature enhancement module, as shown in \hyperref[fig:framework]{Figure~\ref{fig:framework}}, aiming to facilitate the recognition of sketches while accentuating finer details embedded in sketches. Notably, our method exclusively relies on 2D training data, thus circumventing the limitations associated with the lack of 3D training data. Detailed settings are provided in the \textit{Supplementary Material}.
\subsection{Preliminaries}
Since reconstructing 3D face from single sketch is ill-posed, we combine the prior knowledge from FLAME\cite{li2017learning}, a 3D head model that separates the representation of identity shape $\bm{\beta} $, pose parameters $\bm{\theta}$, and expression components $\bm{\psi} $. It can be defined as
\begin{eqnarray}
M\left ( \bm{\beta},\bm{\theta} ,\bm{\psi}  \right ) :\mathbb{R} ^{\left | \bm{\beta} \right |\times \left | \bm{\theta} \right | \times \left | \bm{\psi} \right |} \to \mathbb{R}^{3N},
\end{eqnarray}
where N denotes 5023 vertices. We estimate the FLAME parameters $\bm{\beta}  \in \mathbb{R} ^{100} $, $\bm{\theta}  \in \mathbb{R} ^{6} $, $\bm{\psi}  \in \mathbb{R} ^{50} $, albedo parameters $\bm{\alpha}  \in \mathbb{R} ^{50} $, Spherical Harmonics (SH) lighting $\bm{l}  \in \mathbb{R} ^{27} $, and camera model $\bm{c}  \in \mathbb{R} ^{3} $ for each input sketch. 

Note that $\bm{\alpha}$ helps gain FLAME’s texture map $A\left ( \bm{\alpha} \right ) \in \mathbb{R} ^{d\times d\times 3}$ using an appearance model, which is converted from Basel Face Model’s linear albedo space to the FLAME UV layout. Camera information $\bm{c}$ helps project 3D vertices to the 2D space, consisting of isotropic scale $s  \in \mathbb{R} $ and 2D translation $t  \in \mathbb{R}^{2} $.
Using the differentiable renderer \(R\)\cite{genova2018unsupervised}, we can obtain the rendering result as
\begin{equation}
R\left ( M,\bm{\alpha} ,\bm{l},\bm{c} \right ) \to I_{g},
\end{equation}
where \(I_{g} \) denotes the 2D sketch generated from the 3D shape.
\subsection{Enhancement Module of GCTD}
The feature enhancement module of GCTD (\textbf{G}eometric \textbf{C}ontour and \textbf{T}exture \textbf{D}etail) strengthens sketch features by improving contour detection and refining facial details, ensuring higher-quality representations for subsequent encoders. This non-learning module effectively resolves ambiguity in sketch facial features.

The input sketch is denoised using bilateral filtering, an edge-preserving smoothing method considering spatial proximity $h_s$ and intensity similarity $h_r$. The process is defined as:
\begin{equation}
	g\left ( x,y \right ) = \frac{1}{W}\sum\limits_{a,b}^{} f\left ( x+a,y+b \right ) \cdot h_s\cdot h_r
\end{equation}
where $W$ is the normalization factor, $f\left ( x,y \right )$ represents the processed pixels in the image, $\left ( a,b \right )$ represents the size of the filter. This suppresses noise while preserving prominent facial contours. To further reduce detail loss, the range parameter that governs $h_r$ is adaptively adjusted according to local variance, sharpening edges in high-frequency regions while enhancing smoothing in homogeneous areas. The resulting sketch is cleaner and structurally consistent, enabling more robust feature extraction in subsequent stages.

Subsequently, the sketch is enhanced by balancing the image to make it clearer. The process is defined as
\begin{equation}
	s = T\left ( r \right ) = \int_{0}^{r} p\left ( r \right )dr,
\end{equation}
where $r$ represents the original image grayscale, $s$ represents the transformed image grayscale, and $p\left ( r \right )$ is the probability of the occurrence of $r$. The sketches have more distinct features after being processed by the GCTD module, which is designed with the explicit intention of not producing a realistic image as an intermediate state. The sketches with enhanced features are used for the later facial contour detection, providing a relatively more economical alternative in terms of computational cost.
\subsection{Loss Function}
\paragraph{Weighted Landmark Loss: }This Landmark loss calculates the correspondence of landmarks by compelling the projection of the 3D landmarks to align with the ground-truth landmarks.
\begin{equation}
	L_{lmk}=\sum_{i=1}^{68}\omega_{i} \left \| k_{i}-s\Pi \left ( M_{i} \right ) +t \right \|_{1},
\end{equation}
where \(\Pi\in\mathbb{R}^{2\times3}\) refers to the projection matrix in the process of projecting 3D mesh into the 2D space, and \(M_{i}\in\mathbb{R}^{3}\) is the corresponding vertex of FLAME model M. Different weights \(\omega_{i}\) are assigned to landmarks in the facial regions, with the highest for the jawline and inner mouth, followed by the corners of the nose and mouth, and the lowest for other areas.

\paragraph{Mutual Distance Loss: }Compared to \(L_{lmk}\), the mutual distance loss is less susceptible to misalignment during the reconstruction process and can especially facilitate a more effective capture of expressions and shapes. Simply put, the mutual distance between keypoint pair \(P\) can be defined as
\begin{equation}
	L_{p}=\sum_{\left ( i,j \right ) \in P}^{}\left \| k_{i}-k_{j}-s\Pi \left ( M_{i}-M_{j} \right )  \right \|_{1},
\end{equation}
where \(P\) is a set of keypoint pairs, \(k_{i}\in\mathbb{R}^{2}\) and \(k_{j}\in\mathbb{R}^{2}\) refer to corresponding ground-truth 2D landmarks. Based on this definition, we opt for three categories of keypoint pairs to calculate our mutual distance loss, including keypoint pairs of eyes, inner mouth, and facial contours. To sum up, the loss computes as
\begin{equation}
	L_{md}=\omega_{eyeP}L_{eyeP}+\omega_{mouP}L_{mouP}+\omega_{conP}L_{conP},
\end{equation}
where \(\omega_{x}\) denotes the respective weight.
\paragraph{Photometric Loss: }We use a photometric loss measuring differences between the input sketch \(I_{input}\) and the generated image \(I_{g}\).
\begin{equation}
	L_{pho}=\left \|V_{}\odot \left ( I_{input}-I_{g}\right ) \right \|_{1,1},
\end{equation}
where \(V_{1}\) is a face mask, and the operator \(\odot\) refers to the Hadamard product. Note that this kind of loss is calculated in both the coarse and detail stages to ensure photometric consistency. We designate them as \(L_{phoS}\) and \(L_{phoD}\), respectively, to make a distinction.
\paragraph{Regularization: }
For the coefficients of the coarse and the detail stages, we employ their L2 regularization terms as regularization losses, denoted as \(L_{regS}\) and \(L_{regD}\), respectively.
\paragraph{Overall Loss Functions: }
Overall, in coarse training stage, our optimization goal is
\begin{equation}
	L_{S}=L_{lmk}+L_{md}+\omega_{pho}L_{phoS}+\omega_{reg}L_{regS}.
\end{equation}
In detail training stage, our optimization goal is
\begin{equation}
	L_{D}=L_{md}+\omega_{pho}L_{phoD}+\omega_{reg}L_{regD}.
\end{equation}
The weights of \(L_{pho}\) and \(L_{reg}\) are denoted as \(\omega_{pho}\) and \(\omega_{reg}\), respectively.
\subsection{Domain Adaptation}
Despite incorporating real hand-drawn sketches—such as those from the CUFSF dataset \cite{zhang2011coupled} and our SketchFaces dataset—to enrich the training data, the scarcity of large-scale, high-quality facial sketch datasets still necessitates the use of synthetic data. To address this, we introduce Syn-SketchFaces, a synthetic dataset aimed at increasing both the diversity and volume of training samples.

However, the domain gap between synthetic and real sketches, which stems from differences in texture, stroke patterns, and abstraction levels, poses a significant challenge to the model’s generalization to real-world sketches. In addition, sketches from different sources or artists often exhibit substantial style variations, which further hinder generalization. Inspired by \cite{cai2024single, zhou2023mixstyle}, we incorporate a domain adaptation strategy into our approach, mixing feature statistics from the Syn-SketchFaces dataset and the SketchFaces dataset. We apply the mixing in lower layers, where features are most sensitive to texture, color, and stroke patterns. Let $F_l(\cdot)$ denote the encoder truncated at layer $l$. Given an input sketch $I_{\text{input}}$ and another sketch for mixing $I_{\text{mix}}$, we obtain the feature maps $x = F_l(I_{\text{input}})$ and $x' = F_l(I_{\text{mix}})$.
We then mix their channel-wise statistics at layer $l$ as
\begin{equation}
\label{eq:feature-mix}
\hat{\mathbf{x}}
= \left(\frac{\mathbf{x}-\boldsymbol{\mu}}{\boldsymbol{\sigma}}\right)
  \odot \big(\lambda\,\boldsymbol{\sigma} + (1-\lambda)\,\boldsymbol{\sigma}'\big)
  + \big(\lambda\,\boldsymbol{\mu} + (1-\lambda)\,\boldsymbol{\mu}'\big),
\end{equation}
where $\boldsymbol{\mu},\boldsymbol{\sigma}$ (resp. $\boldsymbol{\mu}',\boldsymbol{\sigma}'$) are the per-channel
mean and standard deviation of $\mathbf{x}$ (resp. $\mathbf{x}'$) computed over spatial
dimensions, $\odot$ denotes channel-wise broadcasting multiplication, 
and $\lambda$ is a per-sample mixing coefficient drawn from a symmetric Beta distribution, controlling the mixing strength. 

This strategy not only bridges the synthetic–real sketch gap but also enhances robustness to intra-domain sketch style variations, resulting in a marked performance gain on diverse sketch inputs.
\begin{figure}[ht]
  \centering
  \begin{subfigure}[b]{0.58\linewidth}
    \includegraphics[width=\linewidth]{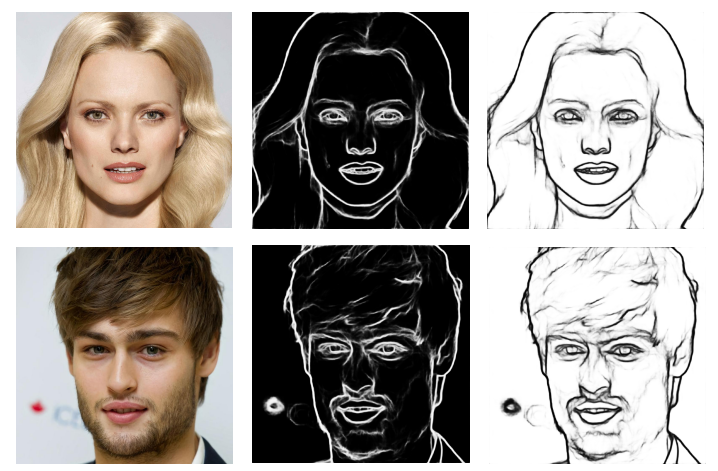}
    \caption{}
    \label{fig:dataset_a}
  \end{subfigure}
  \hfill
  \begin{subfigure}[b]{0.41\linewidth} 
    \includegraphics[width=\linewidth]{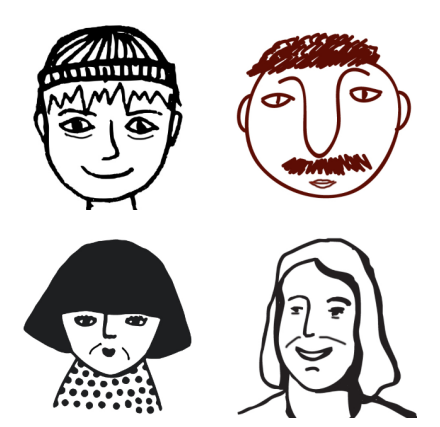}
    \caption{}
    \label{fig:dataset_b}
  \end{subfigure}
  \caption{(a) Syn-SketchFaces generation process: from original photos to PiDiNet contours to synthetic sketches. (b) Real sketch samples from the SketchFaces dataset.}
  \Description{Two subfigures showing datasets used in the study. Subfigure (a) presents the generation process of the synthetic dataset Syn-SketchFaces. It includes three columns: original face photos, contour maps extracted by PiDiNet, and the resulting synthetic sketches. Subfigure (b) shows samples from the SketchFaces dataset, which contains real hand-drawn facial sketches.}
  \label{fig:dataset}
  \vspace{-6pt}
\end{figure}
\begin{figure*}[t]
  \centering
  \includegraphics[width=\linewidth]{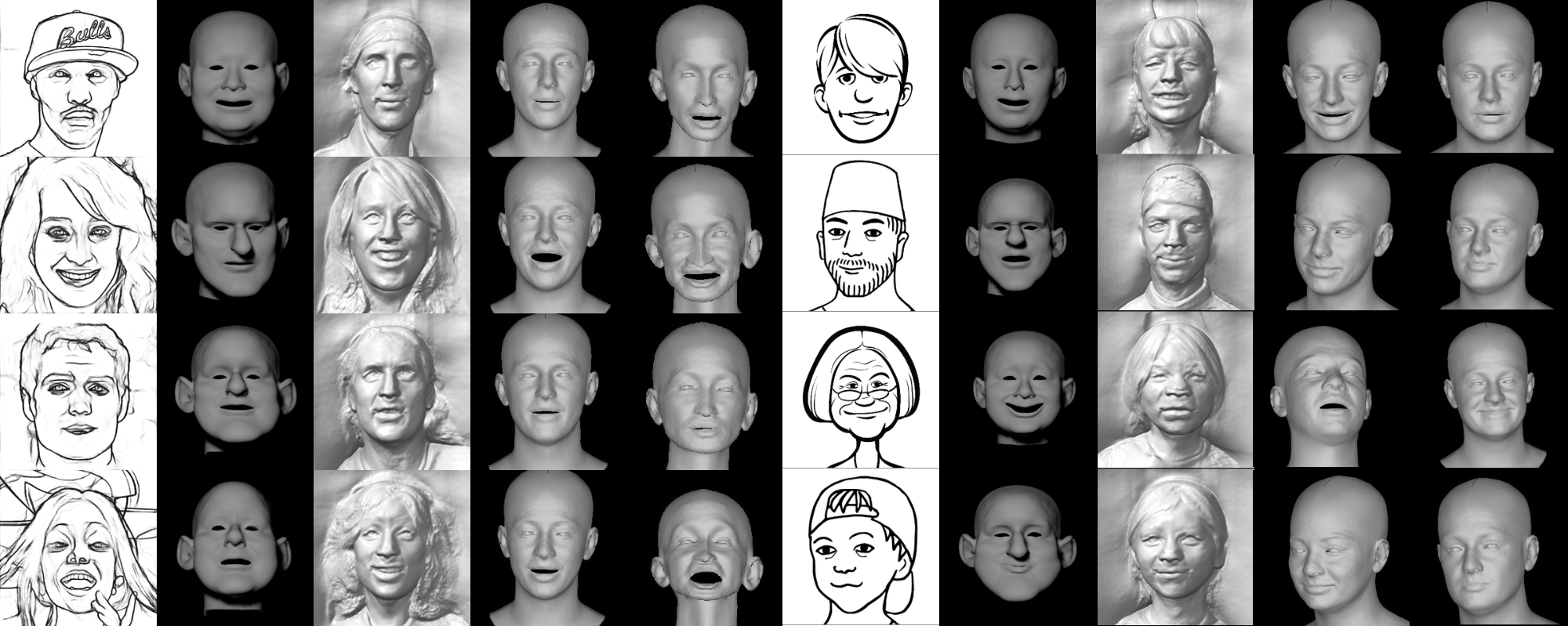}
   \caption{Qualitative comparisons. The left half shows reconstructions from synthetic sketches, and the right half from real hand-drawn sketches. Each group of five columns includes (from left to right): (1) input sketch, (2) DeepSketch2Face\cite{han2017deepsketch2face}, (3) SketchFaceNeRF\cite{lin2023sketchfacenerf}, (4) the combination of\cite{lin2023sketchfacenerf} and\cite{feng2021learning}, and (5) our Sketch-1-to-3.}
   \Description{A qualitative comparison of 3D face reconstruction results on real hand-drawn sketches.}
   \label{fig:comparison}
   \vspace{-6pt}
\end{figure*}
\begin{table*}[tb]
  \caption{Quantitative Comparisons. We employ two metrics, SSIM and GMSD, to evaluate the reconstruction performance of several models. The notation ($\downarrow$) signifies a preference for lower values, while ($\uparrow$) signifies a preference for higher values.}
  \label{tab:Quantitative Comparisons}
   \begin{tabular}{p{2.5cm} p{3cm} p{3cm} p{3cm} p{2cm}}
    \toprule
    \makecell[c]{Method} &
    \makecell[c]{SketchMetaFace\cite{luo2023sketchmetaface}} &
    \makecell[c]{SketchFaceNeRF\cite{lin2023sketchfacenerf}} &
    \makecell[c]{\cite{lin2023sketchfacenerf}-DECA\cite{feng2021learning}} &
    \makecell[c]{Ours} \\
    \midrule
    \makecell[c]{SSIM ($\uparrow$)} &
    \makecell[c]{0.5399} &
    \makecell[c]{0.5209} &
    \makecell[c]{0.9616} &
    \makecell[c]{\textbf{0.9689}} \\
    
    \makecell[c]{GMSD ($\downarrow$)} &
    \makecell[c]{0.5569} &
    \makecell[c]{0.5345} &
    \makecell[c]{0.4584} &
    \makecell[c]{\textbf{0.4072}} \\
    \bottomrule
  \end{tabular}
  \vspace{-6pt}
\end{table*}
\begin{table}[tb]
  \caption{Ablation studies on GCTD and loss functions.}
  \label{tab:ablation}
 \begin{tabular}{p{95pt} p{55pt} p{55pt}}
    \toprule
    \makecell[c]{Model} & \makecell[c]{SSIM ($\uparrow$)} & \makecell[c]{GMSD ($\downarrow$)} \\
    \midrule
    \makecell[c]{Sketch-1-to-3 w/o GCTD} & \makecell[c]{0.9611} & \makecell[c]{0.4469} \\
    \makecell[c]{Sketch-1-to-3 w/o \(L_{conP}\)} & \makecell[c]{0.9676} & \makecell[c]{0.4304} \\
    \makecell[c]{Sketch-1-to-3 w/o \(L_{md}\)} & \makecell[c]{0.9682} & \makecell[c]{0.4265} \\
    \makecell[c]{Sketch-1-to-3 w/o \(L_{lmk}\)} & \makecell[c]{0.9685} & \makecell[c]{0.4259} \\
    \makecell[c]{Sketch-1-to-3} & \makecell[c]{\textbf{0.9689}} & \makecell[c]{\textbf{0.4072}} \\
    \bottomrule
  \end{tabular}
  \vspace{-6pt}
\end{table}
\begin{table*}[tb]
  \caption{User studies. Average ratings (scale: 1–5, poor to excellent) across four aspects for expert (A) and non-expert (B) users.}
  \label{tab:userStudies}
    \begin{tabular}{p{1.5cm} p{3cm} p{3cm} p{3cm} p{3cm}}
    \toprule
    \makecell[c]{Group} &
    \makecell[c]{Modeling Support} &
    \makecell[c]{Convenience} &
    \makecell[c]{Accuracy} &
    \makecell[c]{General Satisfaction} \\
    \midrule
    \makecell[c]{A} & \makecell[c]{4.000} & \makecell[c]{4.267} & \makecell[c]{3.467} & \makecell[c]{3.933} \\
    \makecell[c]{B} & \makecell[c]{4.333} & \makecell[c]{4.400} & \makecell[c]{4.067} & \makecell[c]{4.533} \\
    \bottomrule
  \end{tabular}
  \vspace{-2pt}
\end{table*}
\section{Experiment}
\subsection{Dataset}
There are currently very few facial sketch datasets available, and existing relevant datasets, such as\cite{zhang2011coupled}, can not suffice for training robust sketch-to-3D face reconstruction models because of their limited scales. To optimize the performance of our model, we have collected a real hand-drawn sketch dataset, identified as SketchFaces. Moreover, considering there is a significant cost associated with gathering real hand-drawn sketches, we also generate the synthetic sketch dataset from facial photos. Given the relative abundance of facial photo datasets, we thus obtained a synthetic dataset named Syn-SketchFaces that played a crucial role in the training phase. Furthermore, we use the methods proposed by~\cite{bulat2017far, zou2025towards} to obtain facial landmarks and~\cite{nirkin2018face} for face segmentation, thereby labeling our datasets.
\par
\paragraph{Syn-SketchFaces: }We use PiDiNet~\cite{su2021pixel} to extract gradient cues from images in SCUT-FBP5500~\cite{liang2018scut} and CelebAMask-HQ~\cite{lee2020maskgan}, producing sketch-like results, as shown in \hyperref[fig:dataset_a]{Figure~\ref{fig:dataset_a}}. Each original image is first processed with edge detection, followed by color inversion to generate the final sketches. The resulting dataset preserves both the facial contours and key details.
\par
\paragraph{SketchFaces: }
We gather real hand-drawn facial sketches from individuals with varying levels of artistic proficiency and from a wide range of online sources, capturing diverse sketch styles and drawing mediums, as shown in \hyperref[fig:dataset_b]{Figure~\ref{fig:dataset_b}}. 

\subsection{Quantitative Comparisons}
In Table \ref{tab:Quantitative Comparisons}, we quantify reconstruction results using SSIM, which measures structural similarity, and GMSD, which integrates gradient information to capture edge variations. These metrics jointly provide a comprehensive evaluation. As paired sketch–3D data are unavailable, we project the reconstructed 3D faces into 2D and compare them with the input sketches. All evaluations use the SketchFaces dataset.
\subsection{Qualitative Comparisons}
We conduct qualitative comparisons between our proposed method Sketch-1-to-3 and the current state-of-the-art 3D face reconstruction methods. See \hyperref[fig:comparison]{Figure~\ref{fig:comparison}}. The five columns on the left illustrate the results of the 3D facial reconstruction in synthetic sketches, while the five columns on the right illustrate the results of the 3D facial reconstruction results on real hand-drawn sketches. 

The results of DeepSketch2Face\cite{han2017deepsketch2face} are often abstract and cartoon-like, with limited fidelity to the details of the sketch. In contrast, our Sketch-1-to-3 more accurately captures facial structure, better aligning contours and feature positions. Although SketchFaceNeRF\cite{lin2023sketchfacenerf} appears detailed, many of its reconstructed details do not genuinely originate from the input sketch. Compared to this, our method avoids hallucinated content and faithfully adheres to the sketch input. Furthermore, our method shows greater robustness to occlusions such as hats, while SketchFaceNeRF struggles in such cases.

We also compare our method with a two-step alternative that combines \cite{lin2023sketchfacenerf} and \cite{feng2021learning}, which first synthesizes 2D images from sketches and then reconstructs 3D faces. Although each method excels within its domain, the cross-modal translation, from sketch to photo to 3D, introduces inconsistencies that degrade reconstruction quality. This indirect pipeline tends to hallucinate details absent from the original sketch, making it difficult for the 3D model to capture meaningful feature, as illustrated in \hyperref[fig:sketch-2D-3D]{Figure~\ref{fig:sketch-2D-3D}}. In contrast, our method produces faithful and coherent 3D reconstructions that consistently align well with the input sketches, showing robust performance across diverse sketches. More results across diverse sketch styles are provided in the \textit{Supplementary Material}.
\subsection{Ablation Studies}
\begin{figure}[ht]
  \centering
  \begin{subfigure}[b]{0.48\linewidth}
    \centering
    \includegraphics[width=\linewidth]{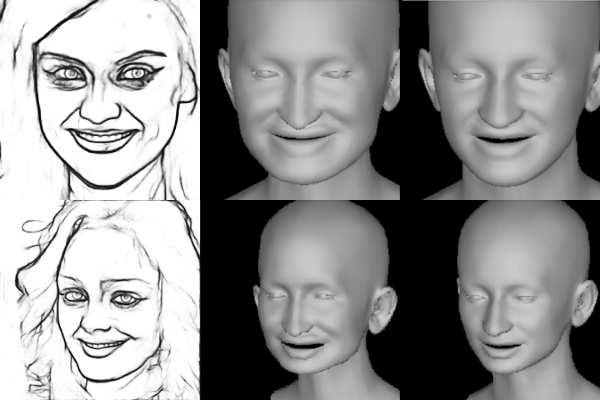}
    \caption{}
    \label{fig:ablation_a}
  \end{subfigure}
  \hfill
  \begin{subfigure}[b]{0.48\linewidth}
    \centering
    \includegraphics[width=\linewidth]{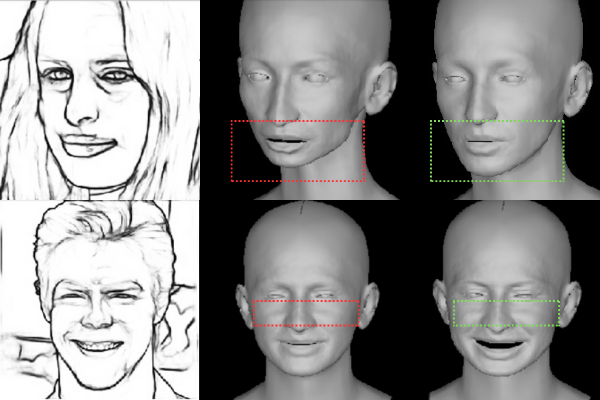}
    \caption{}
    \label{fig:ablation_b}
  \end{subfigure}
  \caption{Ablation studies. In (a), the first row shows the input sketch, Sketch-1-to-3 without \(L_{md}\), and the full model; the second row shows the input sketch, Sketch-1-to-3 without \(L_{conP}\), and the full model. In (b), both rows show the input sketch, Sketch-1-to-3 without GCTD, and the full model.}
  \label{fig:entire_figure}
  \Description{Two subfigures showing results from ablation studies on the Sketch-1-to-3 model. Subfigure (a) includes two rows. The first row shows the input sketch, the reconstruction result without the \(L_{md}\) loss, and the full Sketch-1-to-3 result. The second row shows the input sketch, the result without the \(L_{conP}\) loss, and the full Sketch-1-to-3 result. Subfigure (b) shows two rows, each presenting the input sketch, the result without the GCTD module, and the full model output. These comparisons demonstrate the effectiveness of each component.}
  \vspace{-15pt}
\end{figure}

\hyperref[fig:ablation_a]{Figure~\ref{fig:ablation_a}} and \hyperref[fig:ablation_b]{Figure~\ref{fig:ablation_b}} demonstrate the effectiveness of our proposed loss functions and the GCTD module. Quantitative results in Table~\ref{tab:ablation} further validate their contributions.

\paragraph{Loss Functions: } 
As shown in \hyperref[fig:ablation_a]{Figure~\ref{fig:ablation_a}}, excluding \(L_{md}\) leads to poor chin alignment, while removing \(L_{conP}\) deforms the jawline, highlighting its role in preserving facial structure.
\paragraph{GCTD: }As shown in \hyperref[fig:ablation_b]{Figure~\ref{fig:ablation_b}}, reconstructions with GCTD (third column) exhibit better contour alignment, finer details, and corrected misalignments, demonstrating its effectiveness in enhancing reconstruction accuracy.

\subsection{User Studies}

We conducted a user study to evaluate the real-world applicability of our method with expert users experienced in 3D modeling and non-expert users. Details are provided in the \textit{Supplementary Material}.
\par
The evaluation was conducted along four dimensions: (1) Modeling Support, which assesses the level of assistance provided by the system during the 3D modeling process; (2) Convenience, reflecting the ease and intuitiveness of obtaining results; (3) Accuracy, measuring the perceived correctness and realism of the reconstructed model; (4) General Satisfaction, measuring the overall user experience. As shown in Table\ref{tab:userStudies}, our method received favorable scores from both expert (Group A) and non-expert (Group B) users. Notably, Group B consistently rated the system higher across all evaluation dimensions, with the most significant improvements observed in Accuracy and General Satisfaction. These results suggest that our method is particularly effective in supporting novice users, enabling them to achieve satisfactory 3D face modeling outcomes.

\section{Conclusion}
In this paper, we present Sketch-1-to-3, an end-to-end framework for high-fidelity 3D face reconstruction from a single sketch. Our method addresses the unique challenges posed by sketch-based 3D face reconstruction, including sparse and abstract representations, diverse artistic styles, and the scarcity of sketch data. By introducing the GCTD module, we effectively enhanced the extraction of facial geometry and subtle features from sketches, while our domain adaptation strategy and diverse loss function ensured precise alignment between 2D sketches and 3D reconstruction. 

To further advance research in this domain, we constructed two datasets: SketchFaces and Syn-SketchFaces. Extensive experiments demonstrated that Sketch-1-to-3 achieves state-of-the-art performance, accurately capturing both coarse geometric structure and delicate details, while exhibiting robustness to occlusions, stylistic variations, and incomplete strokes. Further discussion of future work can be found in the \textit{Supplementary Material}.

\balance         

\clearpage

\appendix

\section{Overview}
In this supplementary material, we provide additional details of our Sketch-1-to-3 framework, including implementation details, application scenarios, reconstruction results across diverse sketch styles, additional samples from the SketchFaces dataset, a user study description, and a discussion of limitations and future work.

\section{Implementation Details}
During the first training stage, we conduct training for 10 epochs with a batch size of 16. In the second training stage, we set training for 15 epochs with a reduced batch size of 6. We employed Adam\cite{kingma2014adam} as the optimizer with a learning rate of 1e-4. We set \(\omega_{eyeP}=10.0\), \(\omega_{mouP}=5.0\), \(\omega_{conP}=10.0\), \(\omega_{pho}=0.2\), \(\omega_{reg}=1e-05\). When calculating weighted landmark loss, we assigned weights to the inner mouth and the jawline as 3, the weight for the corners of the nose and mouth is set to 1.5, and the remaining parts are assigned a weight of 1. All three encoders share a ResNet-50 backbone\cite{he2016deep}, and we perform inter-layer feature mixing for domain adaptation. 

\section{Application}
The proposed method offers broad applicability in game development, animation, and digital human creation. By enabling sketch-based 3D face reconstruction, it allows artists to skip labor-intensive modeling steps and focus on creative design, while empowering hobbyists to produce high-quality 3D content with a lower entry barrier than traditional modeling tools.

\begin{figure}[H]
  \centering
  \includegraphics[width=1.0\linewidth]{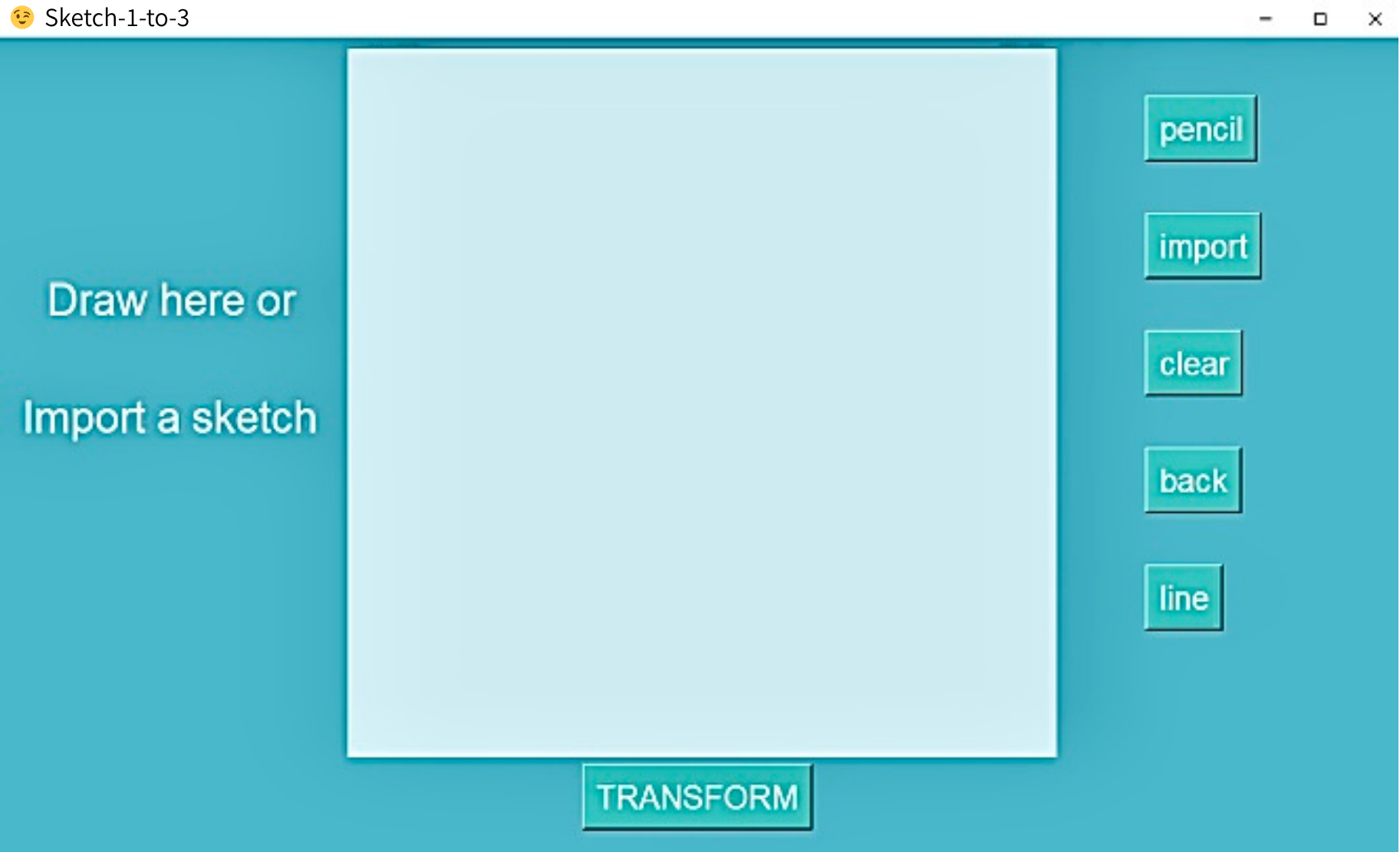}
  \caption{Interface of our system.}
  \Description{Interface of our system.}
  \label{fig:interface}
\end{figure}

\hyperref[fig:interface]{Figure~\ref{fig:interface}} illustrates the interface of our system, which provides an intuitive and flexible sketch-to-3D workflow. The interface supports multiple interactive controls. Users may draw a facial sketch directly within the panel or import an existing one created externally. With a single click, the system automatically generates a detailed and structurally coherent 3D face model from the input sketch, enabling both artists and non-experts to seamlessly transition from 2D concepts to high-quality 3D assets.

\begin{figure}[H]
  \centering
  \includegraphics[width=\linewidth]{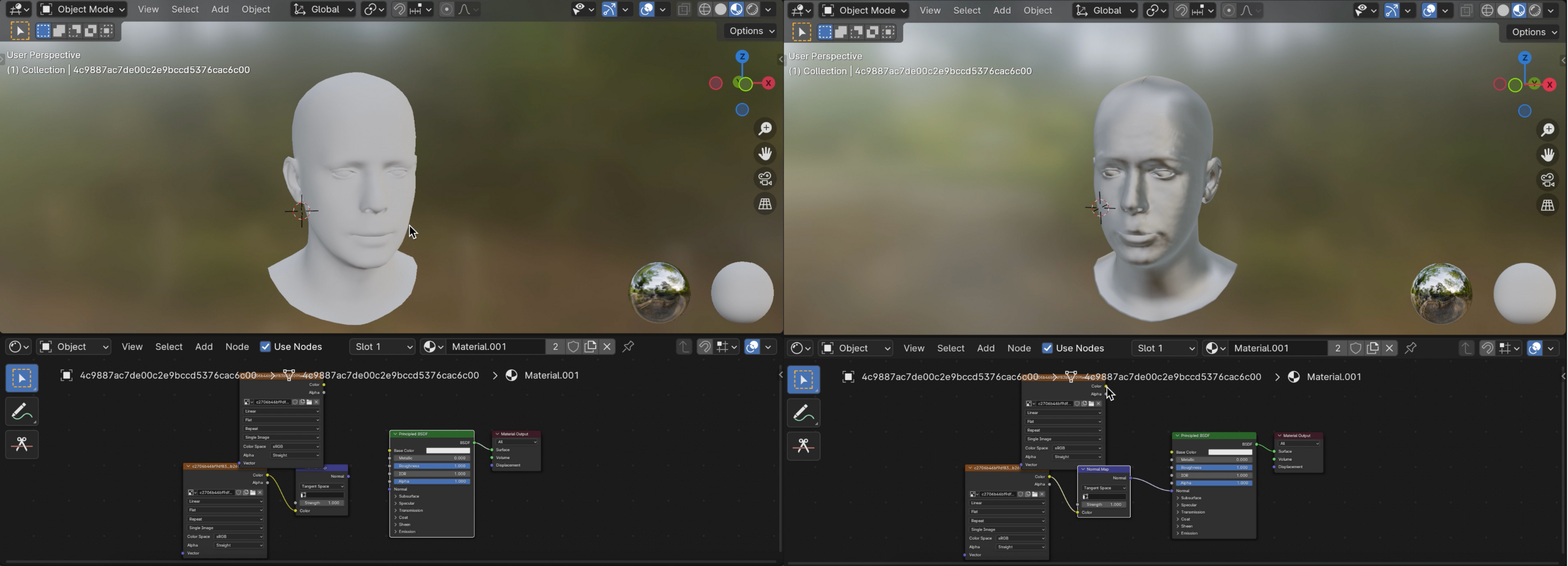}
  \caption{Application in 3D modeling tools.}
  \Description{Application in 3D modeling pipeline.}
  \label{fig:blender}
\end{figure}

\hyperref[fig:blender]{Figure~\ref{fig:blender}} illustrates the application of our method in the 3D modeling pipeline. The reconstructed 3D face can be seamlessly imported into Blender for real-time preview and further editing, supporting a wide range of Blender’s built-in modeling and rendering tools.
\begin{figure*}[t]
    \centering
    \includegraphics[width=1.0\linewidth]{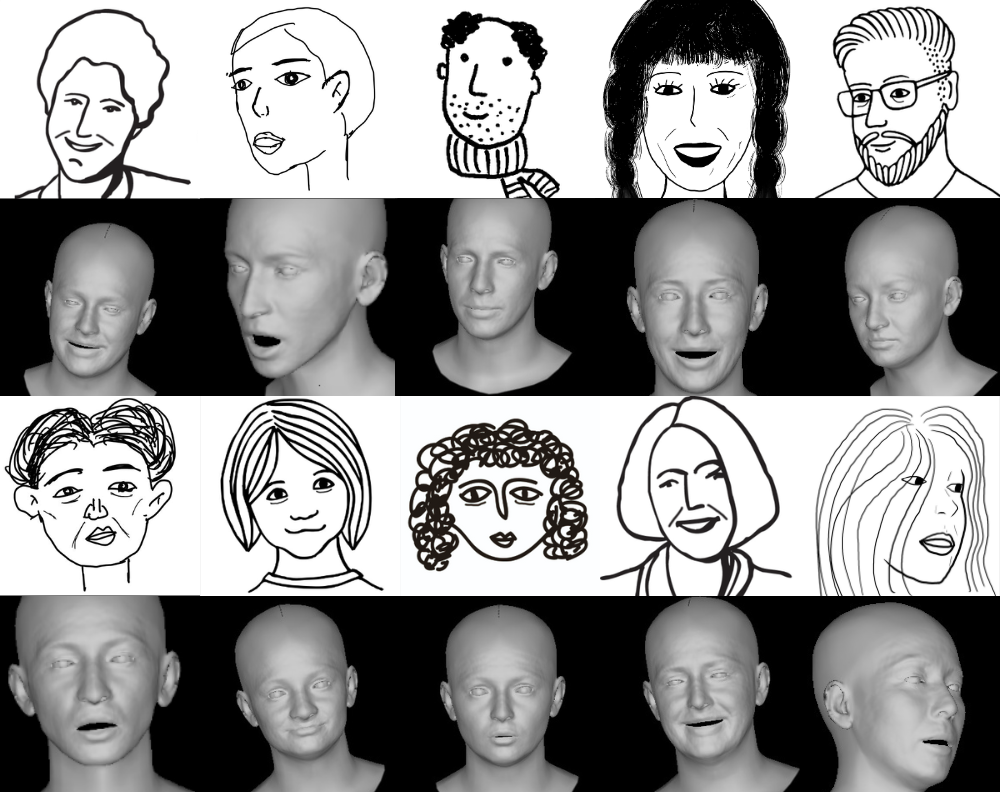}
      \caption{
        More results across various sketch styles. Our method robustly reconstructs 3D faces from sketches with diverse stroke styles, abstraction levels, and viewpoints, showing consistent geometry and stable performance across challenging cases.
        }
      \Description{A two-row image showing 3D face reconstruction from sketches. The top row contains hand-drawn face sketches, and the bottom row shows the corresponding high-fidelity 3D face reconstructions generated by the Sketch-1-to-3 method.}
    \label{fig:more_styles}
\end{figure*}

\begin{figure*}[ht]
  \centering
  \includegraphics[width=1.0\linewidth]{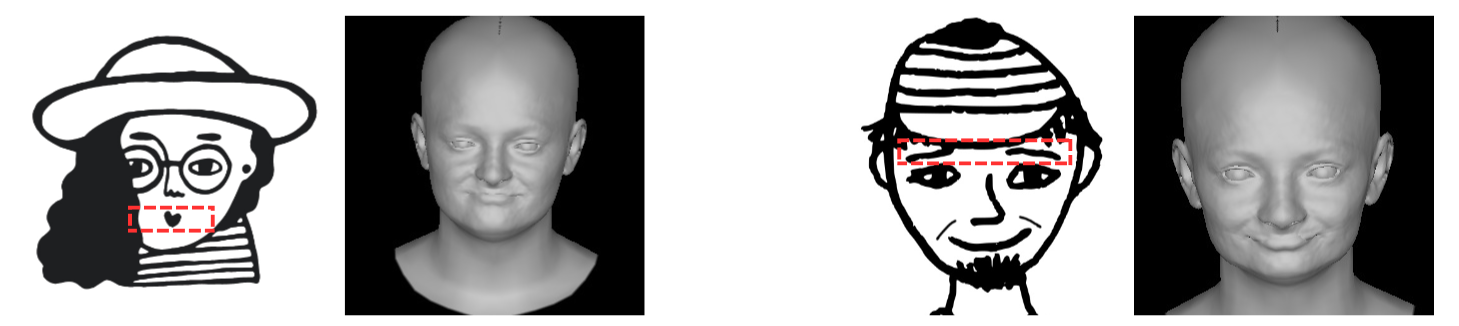}
    \caption{
        Limitation Demonstration. The model faces challenges when interpreting highly abstract or ambiguous sketch features, where complex facial details such as eyebrows or the mouth are reduced to a single dot or line. Such extreme simplifications can lead to uncertainty in geometric reconstruction and highlight the inherent ambiguity of sketch inputs.
    }
  \Description{An illustration showing the model's limitations in interpreting sketch features. Examples highlight cases where simple strokes, such as a dot or a line, are used to represent complex facial details like a mouth or eyebrows, which the model fails to reconstruct accurately.}
  \label{fig:limitation}
\end{figure*}

\section{Experiment}
\subsection{More Results Across Various Sketch Styles}
As shown in \hyperref[fig:more_styles]{Figure~\ref{fig:more_styles}}, we provide additional reconstruction results on sketches of diverse styles, further validating the robustness and generalization capacity of our Sketch-1-to-3 framework. The results show that our model accurately captures facial contours and fine details, while faithfully preserving head orientation, viewpoint, and expression, exhibiting stable performance under occlusions and diverse stroke styles.


\subsection{User Study Design and Procedure}
To assess the real-world applicability of our method, we conducted a user study involving 30 participants (17 males and 13 females) aged between 19 and 26. Participants are divided into two groups based on their 3D modeling experience: Group A consisted of individuals with prior training and practical experience in 3D modeling, while Group B had little to no such experience. Both groups had an equal number of participants.
\par
Each participant was asked to independently draw five sketches of human faces. No specific stylistic restrictions were imposed, allowing participants to express themselves freely. The sketches were encouraged to include facial details, various head orientations, and occlusion to thoroughly test the abilities of our model. After processing the sketches through our model, each participant was shown the corresponding reconstructed 3D face results and asked to rate them on a scale from 1 (poor) to 5 (excellent).

\section{Limitation and Discussion}
Although our method achieves promising results in general situations, there are still areas that can be explored in the future. The sketches inherently incorporate a pronounced level of ambiguity, posing a considerable challenge for the current model to accurately comprehend the intended information in sketches. As depicted in \hyperref[fig:limitation]{Fig~\ref{fig:limitation}}, crucial features such as eyebrows and the mouth of the figure in the sketch can be succinctly represented by a single ink dot or a straight line, which we identify as a key factor contributing to the challenges in sketch-based 3D reconstruction.

Furthermore, as our work focuses on accurate facial geometry and detail reconstruction, we anticipate future efforts to deliver a complete 3D face creation pipeline. This would enable the reconstruction of colorized, texture-mapped 3D faces that faithfully reflect the artist's intent. We think potential solutions to this challenge could come from colorization~\cite{chen2024spcolor}, generation~\cite{zhan2023multimodal}, and more.
\balance 
\clearpage

\bibliographystyle{ACM-Reference-Format}
\bibliography{main}

\end{document}